\documentclass[letterpaper, 10 pt, conference]{ieeeconf}  

\IEEEoverridecommandlockouts                              

\usepackage{graphics} 
\usepackage{epsfig} 
\usepackage{times} 

\usepackage{amsmath,amssymb,amsopn,amstext,amsfonts,amsthm}
\usepackage{pdfsync}
\usepackage{mathtools}
\usepackage{bm}

\newtheorem{definition}{Definition}
\newtheorem{proposition}{Proposition}

\usepackage{algorithm}
\usepackage{algpseudocode}

\usepackage{tabularx}
\usepackage{multirow}
\usepackage{diagbox}
\usepackage{booktabs}
\usepackage{xcolor}

\title{\LARGE \bf
Automated UAV-based Wind Turbine Blade Inspection: Blade Stop Angle Estimation and Blade Detail Prioritized Exposure Adjustment
}

\author{Yichuan Shi$^{1}$, Hao Liu$^{2}$, Haowen Zheng$^{1}$, Haowen Yu$^{1}$, Xianqi Liang$^{1}$, Jie Li$^{2}$, Minmin Ma$^{2}$, and Ximin Lyu$^{1}$%
     \thanks{This work is supported by the Guangdong-Hong Kong-Macao Joint Research of Science and Technology Planning Funding (Grant No. 2023A0505010019) and the National Natural Science Foundation of China (Grant No. 62303495).}
     \thanks{$^1$School of Intelligent Systems Engineering, Sun Yat-sen University, Guangzhou, China. $^2$PowerChina Zhongnan Engineering Corporation Limited, Changsha, China. (\textit{Corresponding author:} Ximin Lyu)}
     \thanks{Email: {\tt\small lvxm6@mail.sysu.edu.cn}}
}

\begin{document}

\maketitle
\thispagestyle{empty}
\pagestyle{empty}

\begin{abstract}

Unmanned aerial vehicles (UAVs) are critical in the automated inspection of wind turbine blades. Nevertheless, several issues persist in this domain. Firstly, existing inspection platforms encounter challenges in meeting the demands of automated inspection tasks and scenarios. Moreover, current blade stop angle estimation methods are vulnerable to environmental factors, restricting their robustness. Additionally, there is an absence of real-time blade detail prioritized exposure adjustment during capture, where lost details cannot be restored through post-optimization. To address these challenges, we introduce a platform and two approaches. Initially, a UAV inspection platform is presented to meet the automated inspection requirements. Subsequently, a Fermat point based blade stop angle estimation approach is introduced, achieving higher precision and success rates. Finally, we propose a blade detail prioritized exposure adjustment approach to ensure appropriate brightness and preserve details during image capture. Extensive tests, comprising over 120 flights across 10 wind turbine models in 5 operational wind farms, validate the effectiveness of the proposed approaches in enhancing inspection autonomy.

\end{abstract}

\section{INTRODUCTION}
\label{sec:intro}

In recent years, propelled by the relentless progress in renewable energy technologies, wind power has emerged as a significant electricity source. As of 2023, the global installed wind power capacity surged to an impressive 1021GW~\cite{GWEC2024wind}. Wind farms strategically dot hills, mountains, and coastal regions, leveraging their bountiful wind resources to offer substantial wind energy potential to support efficient wind turbine installations~\cite{clarke1989wind}. Despite these developments in wind power technology and capacity growth, severe weather conditions such as intense winds, lightning, hail, and rain pose threats to wind turbine blades, potentially leading to damage~\cite{katsaprakakis2021comprehensive}. Regular inspections of these blades are vital for promptly identifying and rectifying any issues, thereby averting disruptions to the turbine's normal operation.

Traditional methods for wind turbine blade inspections have historically relied on manual techniques, involving the use of ground-based telescopes or personnel suspended from turbines using safety ropes. These methods are laborious, inefficient, and pose safety risks~\cite{yang2016progress, dimitrova2022survey}. The introduction of UAV technology has revolutionized these inspections, enabling skilled remote pilots to conduct thorough assessments efficiently~\cite{shafiee2021unmanned}. But it necessitates proficiency in both piloting and photography. The emergence of UAV-based automated inspection systems capitalizes on automation and artificial intelligence, autonomously directing UAV operations during flights and image capture processes~\cite{schafer2016multicopter, parlange2018vision, stokkeland2015autonomous, guo2019detecting, car2020autonomous, castelar2024lidar, yang2024bladeview}. These technological advancements not only mitigate labor costs but also enhance inspection efficiency and elevate image quality.

\begin{figure}[t]
	\vspace{1.0ex}
	\centering
	\includegraphics[width=1.0\columnwidth]{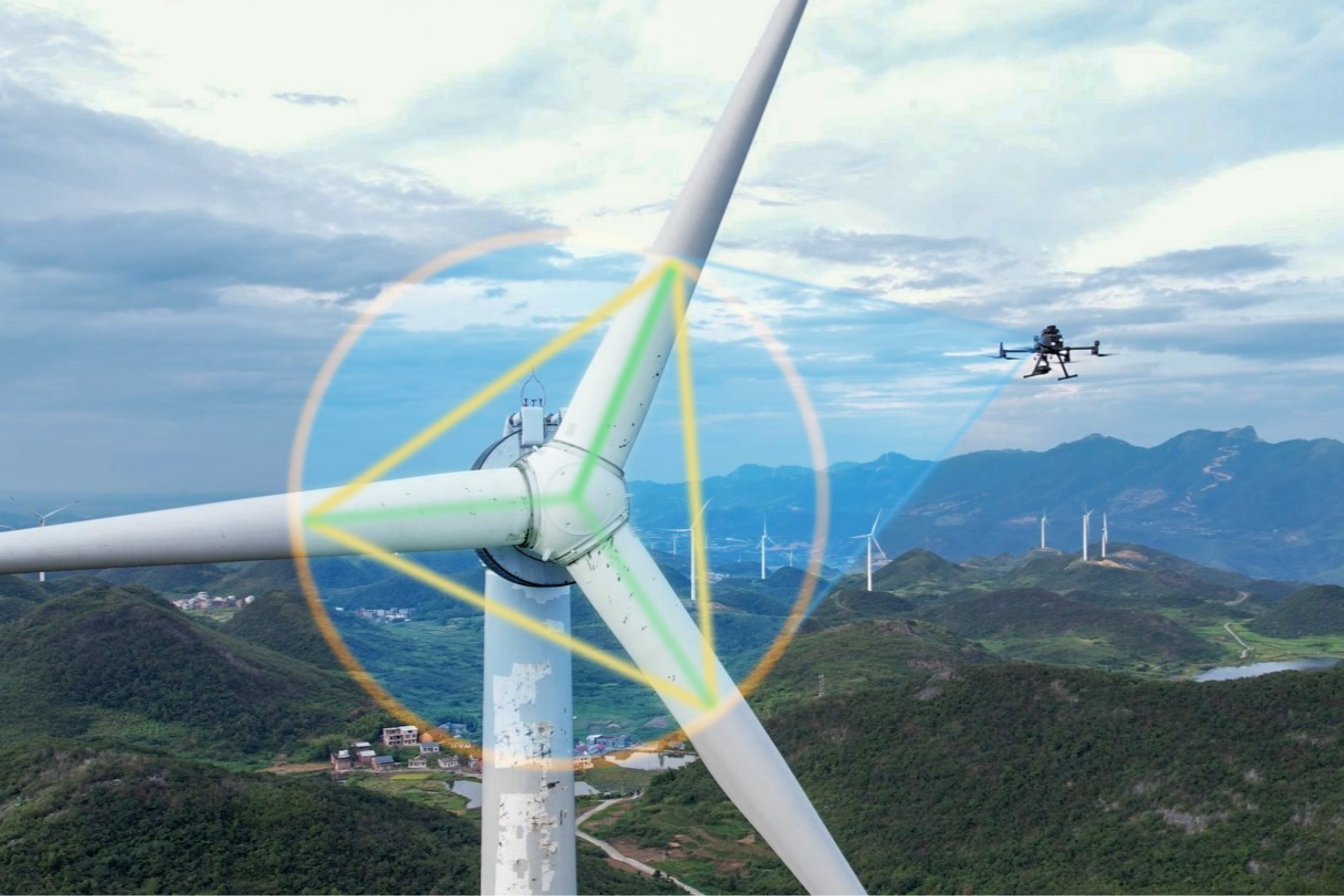}
	\caption{The proposed UAV inspection platform implements the blade stop angle estimation approach in front of a 130m tall wind turbine with 50m long blades in a wind farm in Hunan, China.
		\label{fig:img_cover}}
	\vspace{-2.0ex}
\end{figure}

Nevertheless, despite the extensive research efforts, the field of automated UAV-based wind turbine blade inspection still faces the following challenges.

\subsection{Design of UAV Inspection Platform} 
\label{subsec:intro_platform}

The automated inspection tasks and scenarios impose specific requirements on UAV platform design: 

\textbf{Long Endurance}: Inspections often target large wind turbines, such as those up to 130~$\mathrm{m}$ high with 50~$\mathrm{m}$ blades, necessitating a safety distance of over 10m from the blades, and a flight speed below 2~$\mathrm{m/s}$ to prevent motion blur in image capture. Such long distances and low speeds demand long endurance. 

\textbf{High Wind Resistance}: Wind farms encounter strong wind interference, demanding drones with ample control force and torque, which means large size and high strength, as well as wind resistance algorithm. 

\textbf{Large-scale Perception Capability}: The wide spaces and long blades require extended-range perception capabilities. 

\textbf{High-quality Image Capture Capability}: Subsequent damage detection relies on clarity images with appropriate brightness, underscoring the need for high-quality cameras.

\textbf{High Computing Power}: Implementing automated inspection requires handling substantial perception data, complex navigation and planning algorithms, and ensuring fault tolerance and stability, necessitating robust computing power and memory support.

These demands outlined pose a challenge for the design of UAV inspection platform.

\subsection{Robust Blade Stop Angle Estimation}
\label{subsec:intro_angle}

During inspection, the wind turbine rotor must be stopped by the braking system~\cite{entezami2012fault}. At this point, the angle between the turbine tower and the first blade to its right is defined as the blade stop angle. Given that wind turbine blades can only rotate passively by wind, it is time-consuming and unrealistic to stop the blades at a specified angle, as this would require waiting for the unpredictable wind to position the blades at that angle, especially in windless weather. Hence, the turbine blades may be positioned at any stop angle, resulting in uncertain blade orientation. By estimating the blade stop angle during inspection, it becomes possible to accurately ascertain the blade's position and orientation, thereby aiding in subsequent inspection tasks. This aligns with~\cite{schafer2016multicopter,parlange2018vision}, which highlights that this angle is essential for effective path planning and trajectory generation in wind turbine inspection.

\subsection{Appropriate Blade Detail Prioritized Exposure}
During the inspection, servo control~\cite{car2020autonomous} or trajectory planning~\cite{yang2024bladeview} is utilized for the UAV to fly along the blades and capture images. These images are pivotal for damage detection in subsequent stages. During image capture, variations in sunlight and camera angles may introduce significant brightness disparities between the blade and background regions. However, the automatic exposure adjustment feature integrated into prevalent off-the-shelf cameras prioritize overall image exposure, rather than focusing on specific regions of interest. This can result in overexposure or underexposure in the blade region, potentially obscuring critical details and complicating subsequent damage detection~\cite{yang2023towards,li2023defect}. While manually selecting the blade region on the screen to fine-tune exposure is a workaround, it proves to be impractical for automation purposes. Real-time blade detail prioritized exposure adjustment ensures optimal brightness and enhanced details in the blade region, facilitating subsequent processing.

\vspace{0.5\baselineskip}

To tackle the aforementioned challenges, we introduce a platform and two approaches for automated UAV-based wind turbine blade inspection. Our key contributions are outlined as follows:

\textbf{1) A UAV Inspection Platform}: Addressing the demands of automated inspection tasks and scenarios, we introduce a UAV inspection platform with long Endurance, high wind resistance, large-scale perception capability, high-quality Image capture capability and high computing power.

\textbf{2) A Fermat Point Based Blade Stop Angle Estimation Approach}: Motivated by the unique spatial geometric attributes of wind turbines, we propose a Fermat point based blade stop angle estimation approach that achieves superior accuracy without being affected by background environment.

\textbf{3) A Real-time Blade Details Prioritized Exposure Adjustment Approach}: An exposure adjustment approach is proposed to prioritize the blade detail and adjust the exposure in real-time during capture, ensuring appropriate brightness and detail preservation in varying sunlight conditions.

\textbf{4) Field Validation in Operational Wind Farms}: Extensive testing of over 120 flights across 10 wind turbine models in 5 wind
farms has demonstrated the robust reliability of both the proposed platform and approaches, confirming their efficacy in practical inspection scenarios.

\section{RELATED WORK}
\label{sec:related_work}

\subsection{Automated Wind Turbine Blade UAV Inspection Platform}
Early platforms used for automated wind turbine inspection was utilized for experimental purposes but proved challenging to implement in wind farms. Sch{\"a}fer \textit{et al.}~\cite{schafer2016multicopter} used a Pelican drone equipped with a Hokuyo UTM-30LX LiDAR to conduct inspection in simulation. However, bridging the gap between simulation and reality remains a formidable challenge. Parlange \textit{et al.}~\cite{parlange2018vision} utilize a Parrot Bebop 2 drone with a GoPro camera to carriy out inspection on a small wind turbine model. Nevertheless, the disparity between the small-scale model and the actual wind turbine limits the applicability. Car \textit{et al.}~\cite{car2020autonomous} introduced a custom-build drone equipped with a Velodyne VLP-16 LiDAR. The custom-designed drone's limited body strength and endurance posed challenges in meeting the demands. Recent research has proposed more advanced platforms. Castelar Wembers et al.~\cite{castelar2024lidar} equipped a Hokuyo UST‐20LX LiDAR on a DJI M300 drone, with a perception range limited to a single plane. Yang et al.~\cite{yang2024bladeview} equipped a Velodyne VLP-16 LiDAR on a DJI M600 drone, with a wind resistance of 10m/s. Our proposed platform offers the applicability and stability of a 30min endurance, 12m/s wind resistance, and 70m large-range perception, meeting the demands for inspections.

\subsection{Blade Stop Angle Estimation}
\label{subsec:related_angle}

Certain automated inspection approaches are designed for specific blade stop angles, limiting their applicability~\cite{castelar2024lidar}. Vision-based methods are commonly adopted for blade stop angle estimation. Stokkeland \textit{et al.}~\cite{stokkeland2015autonomous} and Parlange \textit{et al.}~\cite{parlange2018vision} applied traditional vision techniques, utilizing the Hough transform to detect blade lines and determine the blade stop angle. Guo \textit{et al.}~\cite{guo2019detecting} introduced a deep learning based vision approach that identifies the blade tip to ascertain the blade stop angle. Nevertheless, vision-based methods are susceptible to background interferences, potentially causing instability. Yang \textit{et al.}~\cite{yang2024bladeview} proposed a point cloud based method that utilizes geometric relationships and linear fitting. Nonetheless, accuracy challenges may arise when the blade stop angle significantly deviates from 60$^\circ$. Our proposed approach optimizes the hub center and determines the blade stop angle based on the geometric characteristics of wind turbines, achieving higher accuracy and success rates.

\subsection{Blade Region Exposure Adjustment}
\label{subsec:related_exposure}

Certain works employ image enhancement techniques on UAV-taken inspection images to optimize brightness and enhance surface features. Peng \textit{et al.}~\cite{peng2022non} optimized the illumination of the blade region by establishing an illumination model in the Gaussian scale-space of the image. Tan \textit{et al.}~\cite{tan2022research} enhanced the details in the blade region by applying a contrast-limited adaptive histogram equalization algorithm. Nevertheless, restoring details lost during the original capture proves challenging for these post-optimization methods. To the best of our knowledge, there is an absence of relevant research on real-time blade detail prioritize exposure adjustment during inspection capture at present. Our proposed method focuses on retaining the blade detail, adapting the exposure parameter in real time during the original capture, ensuring appropriate brightness levels.

\begin{figure}[t]
	\vspace{2ex}
	\centering
	\includegraphics[width=1.0\columnwidth]{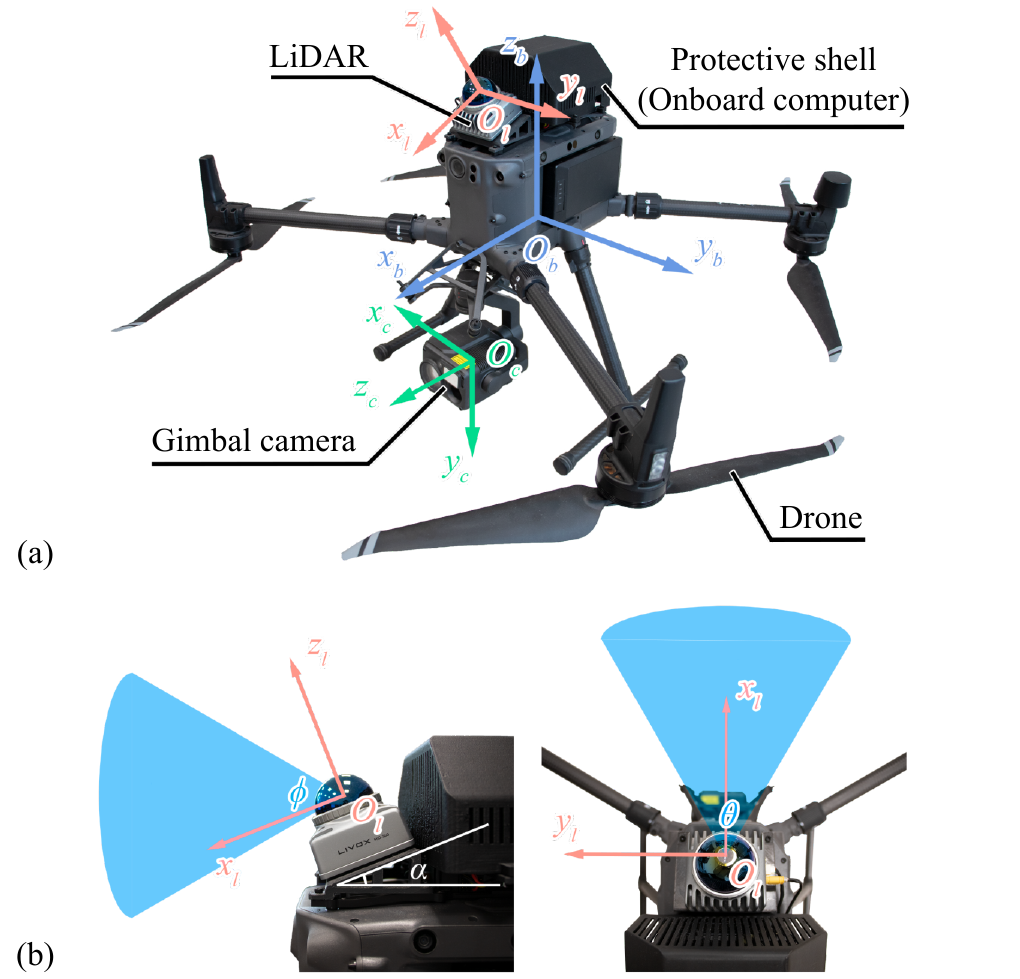}
	\caption{(a) The proposed platform and coordinate system: The drone is equipped with a LiDAR, a gimbal camera and an onboard computer. The body frame (blue) is a FLU frame with the drone's center of gravity as the origin. The LiDAR frame (red) is a FLU frame with the LiDAR's center as the origin. The gimbal frame (green) is a standard camera frame with the image center as the origin. (b) LiDAR Placement: The LiDAR is set up with a pitch angle of $\alpha=-23^\circ$, providing a vertical FOV of $\phi=60^\circ$ and a horizontal FOV of $\theta=60^\circ$ for frontal point cloud perception. 
		\label{fig:img_dronelidar}}
	\vspace{-2ex}
\end{figure}

\section{METHODOLOGY}
\label{sec:methodology}

This section details the methodologies developed to address the three core challenges in Sec.~\ref{sec:intro}: UAV platform design, robust blade stop angle estimation, and real-time blade exposure adjustment.

\subsection{UAV Inspection Platform}
\label{subsec:meth_platform}

To meet the demands from automated inspection tasks and scenarios mentioned in Sec.~\ref{subsec:intro_platform}, we propose a platform depicted in Fig.~\ref{fig:img_dronelidar}(a). The platform is based on the DJI M300 drone, with 30-minute long endurance, 12m/s wind resistance, and obstacle sensing capabilities, ensuring stable and secure inspections. The Livox MID-360 LiDAR, offering a 70m range for 3D perception, is well-suited for the large-scale point cloud acquisition required by perception algorithms. Considering the forward perception demand and the
LiDAR's field of view (FOV), the placement is illustrated in Fig.~\ref{fig:img_dronelidar}(b). The DJI H20T gimbal camera enables 20MP high-quality image capture, camera orientation control and exposure parameter adjustments. The Intel NUC11TNKi5 serves as the onboard computer, with a 4 cores and 8 threads 4.20GHz CPU and 64G memory, providing ample computing power. In addition, a protective shell was designed for the onboard computer using 3D printing to provide protection against rain and sand.

\begin{figure}[t]
	\vspace{2ex}
	\centering
	\includegraphics[width=1.0\columnwidth]{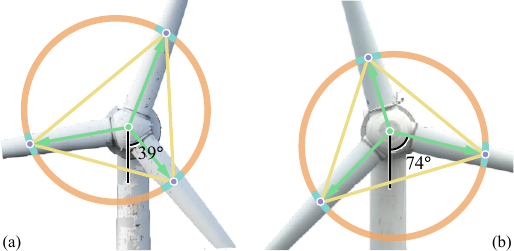}
	\caption{The geometric characteristics of the wind turbine used for blade stop angle estimation. The orange circle represents a ring parallel to the blade rotation plane. The cyan arcs denote the 3 Blade-Ring Intersection (BRI) regions, with the purple points serving as their respective centers, referred to as BRI points. The yellow lines illustrate the BRI triangle formed by the BRI points. The green point indicates the Fermat point of the triangle, also serving as the hub center. The green arrows represent the blade direction vectors formed by connecting the Fermat point and the BRI points. Scenarios (a) and (b) depict different blade stop angles of 39$^\circ$ and 74$^\circ$, respectively.
		\label{fig:img_motivation}}
	\vspace{-4ex}
\end{figure}

\begin{figure*}[t]
	\vspace{2ex}
	\centering
	\includegraphics[width=1.0\textwidth]{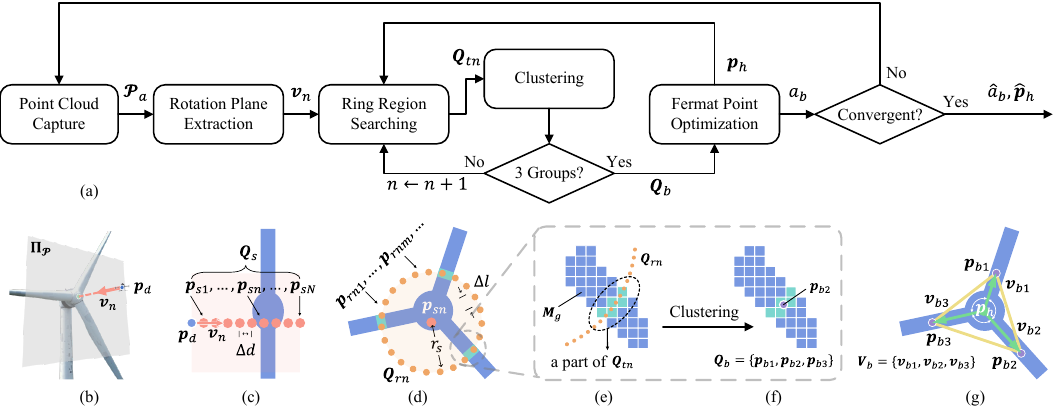}
	\caption{(a) Flowchart of the Fermat point based blade stop angle estimation. (b) The blade rotation plane $\bm{\Pi}_{\bm{\mathcal{P}}}$ (gray plane) and the normal vector $\bm{v}_n$ (red arrow) are extracted. The blue point denotes the drone's position. (c)\&(d) Starting from the drone's position $\bm{p}_d$ (blue point), a series of points $\bm{Q}_s$ (red points) are searched along $\bm{v}_n$ (red arrow) with a step of $\Delta d$. With each $\bm{p}_{sn} \in \bm{Q}_s$ (red point) serving as the center and $r_s$ as the radius, a series of points $\bm{Q}_{rn}$ (orange points) are search with a step of $\Delta l$. (e) Utilizing the probabilistic grid map $\bm{M}_g$ (blue cubes) as the representation of the wind turbine's spatial structure, the turbine blade's spatial segments are searched to obtain the BRI regions $\bm{Q}_{tn}$ (cyan cubes). (f) The BRI regions $\bm{Q}_{tn}$ are clustered into BRI points $\bm{Q}_b$ (purple point). (g) A triangle $\triangle B$ (yellow lines) is formed using $\bm{Q}_b$ and the Fermat point $\bm{p}_h$ (green point) is found. Finally the blade direction vector $\bm{V}_b$ (green arrows) is determined by connecting $\bm{p}_h$ and $\bm{Q}_b$.
		\label{fig:img_angle}}
	\vspace{0ex}
\end{figure*}

\subsection{Fermat Point Based Blade Stop Angle Estimation}
\label{subsec:meth_angle}

Motivated by the unique spatial geometric attributes of wind turbines, we design a blade stop angle estimation approach. This method targets the challenge of environment-sensitive angle estimation mentioned in Sec.~\ref{subsec:related_angle}, leveraging the inherent 120$^\circ$ blade spacing to achieve robust and accurate estimation. The convergence threshold $\varepsilon_a$ is set to ensure the accuracy level required for reliable inspection planning.

As depicted in Fig.~\ref{fig:img_motivation}, envision the utilization of a ring, exceeding the hub's size and parallel to the blade rotation plane, to intersect the 3 blades along the hub’s orientation. This intersection yields 3 Blade-Ring Intersection (BRI) regions on the blade segments, with their respective centers as BRI points. A triangle is formed with these 3 BRI points as the BRI triangle. The Fermat point of this triangle, which defined by Def.~\ref{def:fermat}, can be found according to~\cite{kimberling1994central}.
\begin{definition}
\label{def:fermat}
The Fermat point of a triangle is the geometric median of the three vertices.
\end{definition}
Since the $120^\circ$ between the turbine blades, all interior angles of this triangle are below $120^\circ$. According to Prop.~\ref{prop:fermat}, the Fermat point of the BRI triangle precisely coincides with the hub center.
\begin{proposition}
\label{prop:fermat}
For a triangle with 3 interior angles below $120^\circ$, connecting its Fermat point with the 3 vertices forms 3 line segments, and the angles between them are $120^\circ$.
\end{proposition}
Subsequently, the blade direction vectors can be determined by connecting the Fermat point and the BRI points. 
Following this idea, the proposed approach firstly extracts the blade rotation plane, then obtains the BRI regions through ring region searching. Subsequently, the BRI regions are clustered to obtain the BRI points, and finally utilizes the Prop.~\ref{prop:fermat} to optimize the hub center and computes the blade stop angle. The flowchart of the proposed approach is shown in Fig.~\ref{fig:img_angle}(a). Further details are outlined below.

\subsubsection{Blade Rotation Plane Extraction}
As shown in Fig.~\ref{fig:img_angle}(b), the drone hovers in front of the wind turbine, orienting towards the hub, with the position denoted as $\bm{p}_d$. $\bm{p}_d$ satisfies $\bm{p}_d = \bm{p}_h^0 - \bm{v}_n \cdot d_c$, where $d_c$ is the desired distance between the drone and the hub, and $\bm{p}_h^0$ is a prior estimate of hub position with considerable errors, obtained from~\cite{stokkeland2015autonomous} as the drone ascends along the tower. The point cloud of the wind turbine is captured by LiDAR, denoted as $\bm{\mathcal{P}}_a$. Plane fitting is applied on $\bm{\mathcal{P}}_a$ using the Random Sample Consensus algorithm~\cite{Rusu2011}, resulting in the blade rotation plane $\bm{\Pi}_{\bm{\mathcal{P}}}$ and its unit normal vector $\bm{v}_n$.

\subsubsection{Ring Region Searching and Clustering}
Firstly, the spatial segments of the turbine blades are searched in the ring region mentioned in Fig.~\ref{fig:img_motivation} to obtain the BRI regions, then the BRI regions are clustered to get BRI points. The details of ring region searching is outlined below. As depicted in Fig.~\ref{fig:img_angle}(c), taking $\bm{p}_d$ as the starting point, a series of points $\bm{Q}_s$ are searched along $\bm{v}_n$ with a step of $\Delta d$, as defined in~\eqref{equ:search_point}.
\begin{equation}
\label{equ:search_point}
\bm{Q}_s = \{ \bm{p}_{sn} \in \mathbb{R} ^ 3 \mid \bm{p}_{sn} = \bm{p}_d + n\Delta d\bm{v}_n), n = 1,...,N \}
\end{equation}
where $N$ denotes the max search number. Then, as depicted in Fig.~\ref{fig:img_angle}(d), a plane $\bm{\Pi}_{sn}$ is established on $\bm{p}_{sn}$ with $\bm{v}_n$ as the normal vector. Taking $\bm{p}_{sn}$ as the center and $r_s$ as the radius, a series of points $\bm{Q}_{rn}$ are searched with a step of $\Delta l$, as defined in~\eqref{equ:search_point_r}.
\begin{equation}
\label{equ:search_point_r}
\bm{Q}_{rn} = \left\{ \bm{p}_{rnm} \in \mathbb{R} ^ 3 \ \middle| \ 
\begin{aligned}
& \lVert \bm{p}_{rnm} - \bm{p}_{sn} \rVert_2 = r_s, \\
& \lVert \bm{p}_{rnm} - \bm{p}_{rn(m - 1)} \rVert_2 = \Delta l, \\
& \bm{p}_{rnm} \in\bm{\Pi}_{sn}, m = 1, ..., M
\end{aligned}
\right\}
\end{equation}

The probabilistic grid map is utilized as the spatial representation of wind turbine. The probabilistic grid map~\cite{thrun2002probabilistic} discretizes space at a certain resolution to form regular grid cells. Each grid cell's state is characterized by a probability model: occupied (obstacle), free (passable), or unknown (unexplored). The LiDAR-captured point cloud $\bm{\mathcal{P}}_a$ is fused into the grid map $\bm{M}_{g}$, transforming the wind turbine's spatial structure from an unordered point cloud to an ordered grid map representation. As depicted in Fig.~\ref{fig:img_angle}(e), during the ring region searching, if the searching point $\bm{p}_{rnm}$ falls within an occupied grid cell, the grid cell is recognized as part of the BRI regions $\bm{Q}_{tn}$. We select the grid map representation over the point cloud due to two primary advantages: (1) Grid map query operations have a time complexity of $\mathcal{O}(1)$, which is more efficient than nearest neighbor search operations on point cloud optimized using KD-Tree, with a complexity of $\mathcal{O}(\log n)$, ensuring effective computations for large-scale wind turbines. (2) It is compatible with grid map based trajectory planning in future works without introducing excess computational overhead.

Subsequently, clustering is performed on the BRI regions $\bm{Q}_{tn}$ to acquire the BRI points $\bm{Q}_{b}$, as depicted in Fig.\ref{fig:img_angle}(f). The DBSCAN algorithm\cite{ester1996density} is utilized for clustering due to its ability to adapt to various cluster numbers and noise robustness. Successful clustering involves dividing into 3 groups, each denoting a BRI region on a blade. If clustering fails, the searching is repeated at $\bm{Q}_{s(n + 1)}$ until successful clustering is achieved. Following clustering, the cluster centers of the 3 groups represent the 3 BRI points $\bm{Q}_{b} = \{ \bm{p}_{b1}, \bm{p}_{b2}, \bm{p}_{b3} \}$.

\begin{algorithm}[h]
\caption{Fermat point based blade stop angle estimation}
\label{alg:agldet}
\vspace*{0.5ex}
\hspace*{0.02in} {\bf Input:}
Initial hub position $\bm{p}_h^0$, Convergence threshold $\varepsilon_a$. \\
\hspace*{0.02in} {\bf Output:}
Blade stop angle $\hat{a}_b$, Hub center $\hat{\bm{p}}_h$
\begin{algorithmic}[1]
\For{iteration $i$}
\State Capture Point cloud $\bm{\mathcal{P}}_a^i$ by LiDAR.
\State Fit $\bm{\mathcal{P}}_a^i$ into plane $\bm{\Pi}_{\bm{\mathcal{P}}}^i$ and get vector $\bm{v}_n^i$.
\State Get search points $\bm{Q}_s^i$ with $\bm{p}_h^{i-1}$ by~\eqref{equ:search_point}.
\For{each $\bm{p}_{sn}^i$ in $\bm{Q}_s^i$}
\State Get search points $\bm{Q}_{rn}^i$ by~\eqref{equ:search_point_r}. 
\State Search all $\bm{p}_{rnm}^i \in \bm{Q}_{rn}^i$ in $\bm{M}_g$ and get $\bm{Q}_{tn}^i$.
\State Cluster $\bm{Q}_{tn}^i$ into $k$ groups.
\If{$k$ = 3}
\State Form triangle $\triangle B^i$ by cluster centers $\bm{Q}_b^i$.
\State Find Fermat point $\bm{p}_h^i$ of $\triangle B^i$.
\State Compute blade stop angle $a_b^i$.
\State \textbf{break}.
\EndIf
\EndFor
\If{$| a_b^i - a_b^{i-1} | < \varepsilon_a$}
\State $\hat{a}_b \leftarrow a_b^i$. $\hat{\bm{p}}_h \leftarrow \bm{p}_h^i$.
\State \textbf{return} $\hat{a}_b$ and $\hat{\bm{p}}_h$.
\EndIf
\EndFor
\end{algorithmic}
\end{algorithm}

\subsubsection{Fermat Point Optimization}
Finally, the hub center is optimized utilizing the Prop.~\ref{prop:fermat} and the blade stop angle is computed, as shown in Fig.~\ref{fig:img_angle}(g). A triangle $\triangle B$ is constructed using $\bm{Q}_b$. The Fermat point of $\triangle B$ is determined following~\cite{kimberling1994central} and is applied as the optimal estimation for the hub center $\bm{p}_h$. The hub center $\bm{p}_h$ and $\bm{Q}_b$ is connected to determine the blade direction vector $\bm{V}_b = \{ \bm{v}_{b1}, \bm{v}_{b2}, \bm{v}_{b3} \}$ and compute the blade stop angle $a_b$. The algorithm's convergence criterion is met when $a_b^i$ obtained in the $i$-iteration and $a_b^{i-1}$ acquired in the $(i - 1)$-iteration satisfy $| a_b^i - a_b^{i - 1} | < \varepsilon_a$, leading to the optimal $\hat{a}_b$ and $\hat{\bm{p}}_h$. The $\varepsilon_a$ denotes the convergence threshold. The algorithm procedure is summarized in Algorithm.~\ref{alg:agldet}.

\subsection{Blade Details Prioritized Exposure Adjustment}
\label{subsec:meth_exposure}

To ensure clear details of the blade region in the captured images, we introduce a blade detail prioritized exposure adjustment approach. This approach addresses the gap in real-time blade-focused exposure mentioned in Sec.~\ref{subsec:related_exposure}, adjusting the camera exposure parameter to keep blade surface illumination within $ \left[ \mu_{\text{min}}, \mu_{\text{max}}\right] $, which is derived from the specific requirements of subsequent damage detection tasks.

Considering that the point cloud provides a spatial representation of the blade, the proposed approach initially utilizes the point cloud to extract the current inspection point on the blade. Subsequently, this inspection point is projected onto the image, serving as the reference point for exposure adjustment. Finally, the camera exposure parameter is adjusted according to the brightness within the reference region surrounding the reference point. Further details are outlined below. 

\begin{figure}[t]
	\vspace{2ex}
	\centering
	\includegraphics[width=\columnwidth]{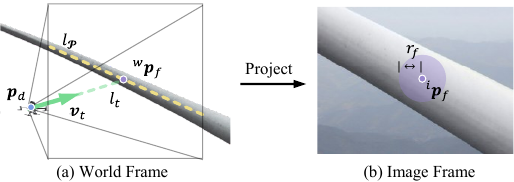}
	\caption{Blade detail prioritized exposure adjustment. (a) $\bm{p}_d$ is the drone's position. The yellow and green dotted lines represent the blade center line $l_{\bm{\mathcal{P}}}$ and the perpendicular line $l_t$, respectively. The green arrow represents the perpendicular vector $\bm{v}_t$, and the purple point indicates the current inspection point ${}^{w}\bm{p}_{f}$. (b) The purple point indicates the reference point for exposure adjustment ${}^{i}\bm{p}_{f}$, and the purple circle is the reference region $\bm{C}_f$.
		\label{fig:img_track}}
	\vspace{-2ex}
\end{figure}

Initially, the current inspection point on the blade is extracted. Considering that capturing images perpendicular to the blade surface yields optimal imaging results, we leverage geometric relationships to determine the perpendicular foot as the current inspection point. As illustrated in Fig.~\ref{fig:img_track}, with the blade's distinctive shape, line fitting is applied on the LiDAR-captured blade point cloud $\bm{\mathcal{P}}_b$ using the RANSAC algorithm~\cite{Rusu2011} , resulting in the blade center line $l_{\bm{\mathcal{P}}}$. A perpendicular line $l_t$ is established from the drone's position $\bm{p}_d$ to $l_{\bm{\mathcal{P}}}$, with ${}^{w}\bm{p}_{f}$ as the perpendicular foot, satisfying~\eqref{equ:perp3}.
\begin{equation}
\label{equ:perp3}
\bm{p}_d \in l_t, l_t \perp l_{\bm{\mathcal{P}}}, l_t \cap l_{\bm{\mathcal{P}}} = {}^{w}\bm{p}_{f}
\end{equation}
The perpendicular foot ${}^{w}\bm{p}_{f}$ on the blade center line is served as the current inspection point. The gimbal is adjusted to align with the $l_t$'s direction vector $\bm{v}_t$, ensuring that the blade center remains centered in the captured image.

Subsequently, the reference point for exposure adjustment is obtained by projecting. Following the principle of perspective projection~\cite{faugeras1993three}, the current inspection point ${}^{w}\bm{p}_{f}\in\mathbb{R}^3$ in the world frame is projected into the image frame, denoted as $^{i}\bm{p}_f\in\mathbb{R}^2$, refer to~\eqref{equ:proj3to2}.
\begin{equation}
\label{equ:proj3to2}
z_c \begin{bmatrix} {}^{i}\bm{p}_{f} \\ 1 \end{bmatrix} = \bm{K}_c \cdot {}^{c}\bm{p}_{f} = \bm{K}_c \cdot \begin{bmatrix} {}^{c}\bm{R}_{w} & {}^{c}\bm{t}_{w} \end{bmatrix} \cdot \begin{bmatrix} {}^{w}\bm{p}_{f} \\ 1 \end{bmatrix}
\end{equation}
where $\begin{bmatrix} {}^{c}\bm{R}_{w} & {}^{c}\bm{t}_{w} \end{bmatrix}\in\mathbb{R}^{3 \times 4}$ is the extrinsic matrix, representing the transformation from the world frame to the camera frame. ${}^{c}\bm{p}_{f}$ represents the inspection point in the camera frame, and $z_c$ is the z-component of ${}^{c}\bm{p}_{f}$. $\bm{K}_c\in\mathbb{R}^{3 \times 3}$ is the camera's intrinsic matrix. Following projection, the projected current inspection point lies in the blade center, serving as the reference point of exposure adjustment.

Finally, the camera exposure parameter is adjusted based on the brightness of the reference region. The blade surface's exposure is assessed through the brightness of the region surrounding ${}^{i}\bm{p}_{f}$. Specifically, a circular region $\bm{C}_f$ centered at ${}^{i}\bm{p}_{f}$ is defined as the reference region, and the grayscale level within this region is computed, as depicted in~\eqref{equ:area_circle}:
\begin{equation}
\label{equ:area_circle}
\bm{C}_f = \{ {}^{i}\bm{p} \mid {}^{i}\bm{p} \in \bm{G}, \lVert {}^{i}\bm{p} - {}^{i}\bm{p}_{f} \rVert_2 \leq r_f \}
\end{equation}
where $\bm{G}$ represents the grayscale image converted from the real-time gimbal camera image, and $r_f$ denotes the region's radius. Subsequently, the mean grayscale value $\mu_g$ within $\bm{C}_f$ is computed as shown in~\eqref{equ:area_aver}:
\begin{equation}
\label{equ:area_aver}
\mu_g = \dfrac{1}{|\bm{C}_f|} \sum\nolimits_{{}^{i}\bm{p} \in \bm{C}_f} \bm{G}({}^{i}\bm{p})
\end{equation}
where, $|\bm{C}_f|$ indicates the number of pixels within $\bm{C}_f$. $\mu_g$ is used to represent the brightness level of the blade region. To ensure that $\mu_g$ remains within the appropriate range $ \left[ \mu_{\text{min}}, \mu_{\text{max}}\right] $, the current exposure parameter $\gamma$ is adjusted with a step factor $k_\mu$. The exposure adjustment algorithm is outlined in Algorithm.~\ref{alg:evadj}. 

\begin{algorithm}[b]
\caption{Blade detail prioritized exposure adjustment}
\label{alg:evadj}
\vspace*{0.5ex}
\hspace*{0.02in} {\bf Input:}
Grayscale range $[\mu_{\text{min}}, \mu_{\text{max}}]$, Adjust factor $k_\mu$.
\begin{algorithmic}[1]
\While {inspecting}
\State Capture Point cloud $\bm{\mathcal{P}}_b$ by LiDAR.
\State Fit $\bm{\mathcal{P}}_b$ to get center line $l_{\bm{\mathcal{P}}}$.
\State Establish line $l_t$ and find ${}^{w}\bm{p}_{f}$ that satisfies~\eqref{equ:perp3}.
\State Project ${}^{w}\bm{p}_{f}$ onto image to get $^{i}\bm{p}_f$ by~\eqref{equ:proj3to2}.
\State Construct circle $\bm{C}_f$ with $^{i}\bm{p}_f$ and $r_f$ on $\bm{G}$ by~\eqref{equ:area_circle}.
\State Calculate mean grayscale value $\mu_g$ by~\eqref{equ:area_aver}.
\If {$\mu_g > \mu_{\text{max}}$}
    \State $\gamma$ $\gets$ $\gamma$ - $k_\mu$;
\ElsIf {$\mu_g < \mu_{\text{min}}$}
    \State $\gamma$ $\gets$ $\gamma$ + $k_\mu$;
\EndIf
\State Set camera exposure value to $\gamma$.
\EndWhile
\end{algorithmic}
\end{algorithm}

\section{EXPERIMENT}
\label{sec:experiment}

We conducted experiments in operational wind farms. To date, the proposed platform and approaches have successfully completed 120 flights involving 10 distinct models of wind turbines across 5 operational wind farms. Throughout these experiments, temperatures varied from 5$^\circ$C to 33$^\circ$C, with maximum wind speeds reaching up to $12$m/s. Despite an occasional interruption caused by sudden light rain, the platform safely returned without any damage. These results highlight the performance of our platform in coping with the complex environment of the wind farm. In the following, we will present the results of the blade stop angle estimation and the exposure adjustment approaches, respectively. 

\subsection{Fermat Point Based Blade Stop Angle Estimation}
\label{subsec:blade_angle_detect_exp}

\begin{figure}[t]
	\vspace{2.0ex}
	\centering
	\includegraphics[width=1.0\columnwidth]{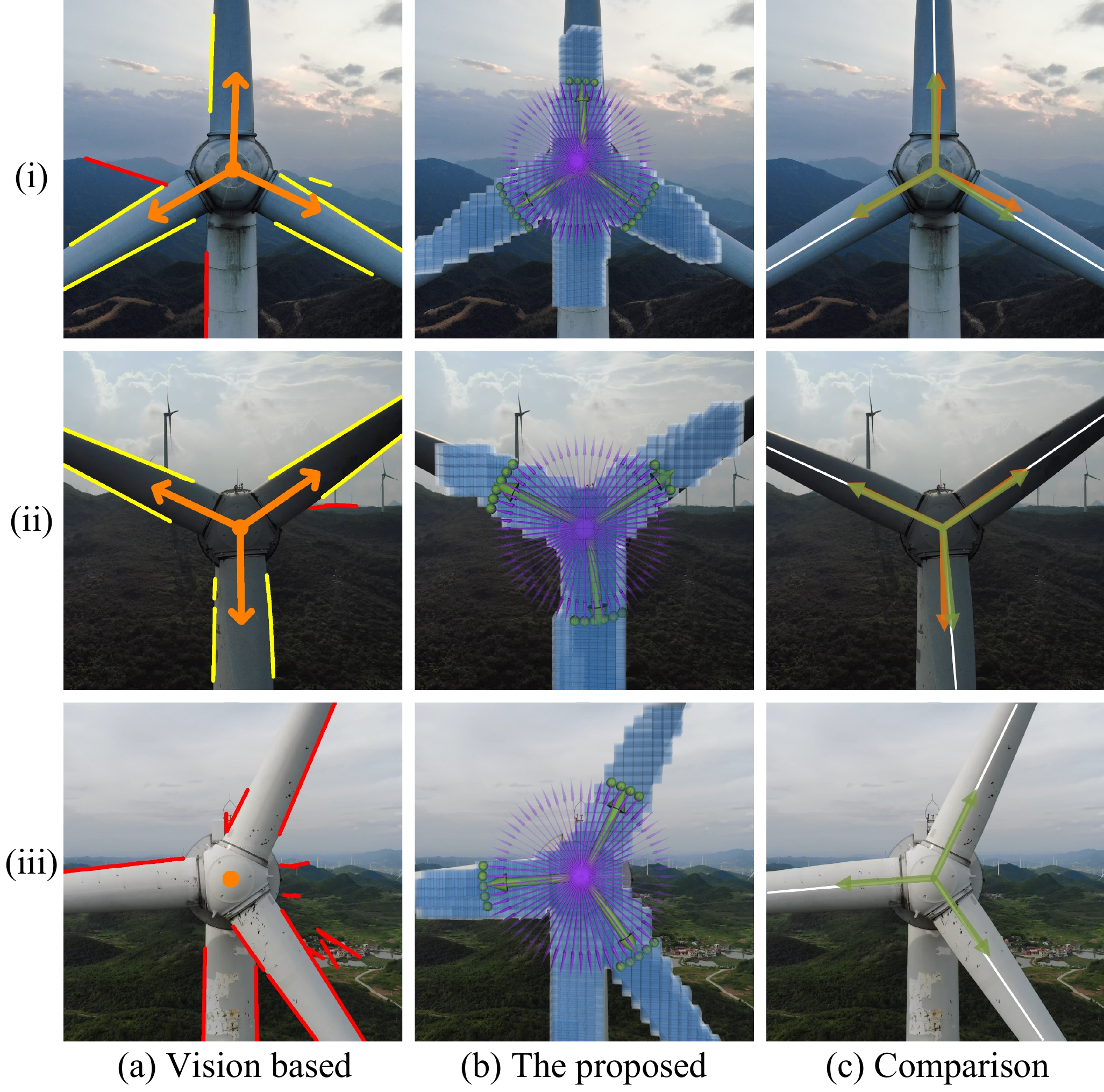}
	\caption{Results of the blade stop angle estimation. (a) The result of the vision-based method. The yellow and red lines are the detected edge lines, representing the blade and non-blade lines recognized by the voting algorithm, respectively. The orange arrows indicate the blade direction vectors. (b) The result of the proposed method. The blue cubes are the spatial representation of the wind turbine. The purple arrows symbolize the ring region searching. The green points represent the BRI regions, and the green arrows denote the blade direction vectors. (c) A comparison of the two methods. The white lines represent the ground truth. The orange and green arrows reflect the results of the vision-based method and the proposed method, respectively.
		\label{fig:img_angle_result}}
	\vspace{-6.0ex}
\end{figure}

\renewcommand{\arraystretch}{1.2}
\begin{table}[b]
\centering
\vspace{0ex}
\caption{Comparison of the mean angle error and success rate}
\label{tab:angle_result}
\vspace{0ex}
\begin{tabular}{c c c} 
\toprule
 & Vision-based & The proposed \\
\midrule
Mean Angle Error $(^{\circ})$ $\bm{\downarrow}$ & 2.14 & \textbf{1.15} \\
Success Rate $\bm{\uparrow}$ & 69.2\% & \textbf{98.3\%} \\
\bottomrule
\end{tabular}
\end{table}

\begin{figure*}[t]
	\vspace{2.0ex}
	\centering
	\includegraphics[width=1.0\textwidth]{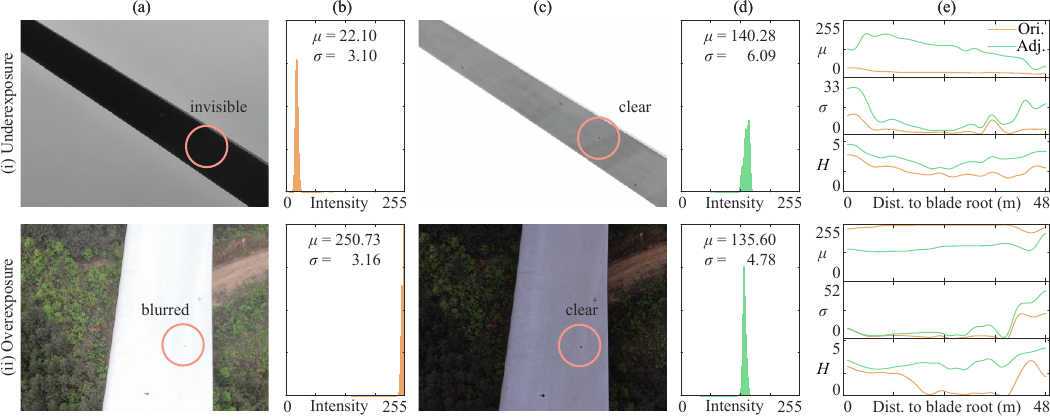}
	\caption{Results of the blade exposure adjustment algorithm, showing two scenarios: (i) underexposure and (ii) overexposure. For both scenarios, columns (a) and (c) display the original and adjusted images, respectively, with their corresponding histograms in columns (b) and (d). Column (e) presents metric curves comparing the original and adjusted images across the inspection process of a single blade surface, where the x-axis represents the distance from the drone to the blade root. (Abbreviation: Ori.: Original, Adj.: Adjusted, Dist.: Distance)
		\label{fig:img_adjust_result}}
	\vspace{0ex}
\end{figure*}

Throughout each test, the wind turbine's braking system was activated to stop the turbine rotor, with the diverse wind conditions on site naturally yielding varying blade stop angles. We compared our experimental findings with the established vision-based method. Stokkeland \textit{et al.}~\cite{stokkeland2015autonomous} proposed a vision-based method, employing the Hough line transform~\cite{duda1972use} to detect lines and a voting algorithm to identify blade lines. This method was adapted to suit our estimation distance while preserving its crucial voting algorithm. The experimental results are illustrated in Fig.\ref{fig:img_angle_result}. Fig.\ref{fig:img_angle_result}(i)\&(ii) show the interference of the background mountains and wind turbine tower to the visual-based method respectively, leading to inaccuracies in identifying the right and lower blades, consequently affecting the estimation results. In Fig.~\ref{fig:img_angle_result}(iii), the intricate visual details of the background village render the voting algorithm in the vision-based method unable to recognize the turbine blades, resulting in a estimation failure. In contrast, the proposed method utilizes accurate point cloud from LiDAR, searches for the turbine blades based on the blade rotation plane, remaining unaffected by background visual factors and the turbine tower.

Tab.~\ref{tab:angle_result} compares the mean angle error and success rates of the two methods. The proposed method outperforms the vision-based method, with a lower mean angle error (1.15$^{\circ}$ vs. 2.14$^{\circ}$) and a significantly higher success rate (98.3\% vs. 69.2\%). These findings underscore the superior robustness and accuracy of the proposed method. Among these 120 trials, 2 instances of failure were observed. One arose from heavy fog, which impaired LiDAR’s perception of blade structure. The other stemmed from large errors in the prior hub position estimate $\bm{p}_h^0$, leaving the ring region searching unable to extract the BRI regions. Future work will focus on improving the accuracy of the prior hub position estimate.

\subsection{Blade Details Prioritized Exposure Adjustment}
\label{subsec:blade_exposure_adjust_exp}

The experiments were conducted on sunny, cloudy, and overcast days to introduce varying sunlight conditions for testing the approach in both underexposed and overexposed scenarios. The proposed platform was controlled to inspect the turbine blade, capturing images at predefined intervals from the blade root to the blade tip. The mean $\mu$, standard deviation $\sigma$, and entropy $H$ of the gray value within the blade region serve as metrics to evaluate our method~\cite{shirvaikar2004optimal}. Specifically, $\mu$ reflects brightness, and $\sigma$ and $H$ signify the intricacy of details. Typically, underexposure is prevalent on the lower blade surface, and overexposure is common on the upper surface. The evaluation of these two scenarios is detailed in Fig.~\ref{fig:img_adjust_result}.

Comparing the original and adjusted images, we analyze their histograms within the blade region. The position and width of the histogram reflect brightness and detail richness, respectively. Underexposure, as shown in Fig.~\ref{fig:img_adjust_result}(i)-(a), manifests as dimness and invisible details (red circle in Fig.~\ref{fig:img_adjust_result}(i)-(a)). The adjusted image Fig.~\ref{fig:img_adjust_result}(i)-(c) enhances brightness to reveal clearer details. In Fig.~\ref{fig:img_adjust_result}(i)-(b) and (i)-(d), the corresponding histograms show a transition from a low gray value of 22.10 to 140.28, expanding the gray range by 97.0\%. Conversely, overexposure, depicted in Fig.~\ref{fig:img_adjust_result}(ii)-(a), leads to excessive brightness and blurry details (red circle in Fig.~\ref{fig:img_adjust_result}(ii)-(a)). The adjusted image Fig.~\ref{fig:img_adjust_result}(ii)-(c) rectifies this to enhance intricacy of details. In Fig.~\ref{fig:img_adjust_result}(ii)-(b) and (ii)-(d), their corresponding histograms shift from a high gray value of 250.73 to of 135.60, expanding the gray range by 51.3\%.

We compute these metrics for both the original and adjusted images, plotting their curves as the drone inspects from blade root to tip. Fig.~\ref{fig:img_adjust_result}(i)-(e) and (ii)-(e) illustrate underexposed and overexposed scenarios, respectively. The original $\mu$ curve lies at the extremes, while the adjusted curve settles within an optimal range, indicating appropriately adjusted brightness. The adjusted $\sigma$ and $H$ curves surpass the original, signifying richer blade details. Tab.~\ref{tab:exposure_result} showcases the metric means. In underexposed scenarios, the brightness increased from 25.48 to 133.27, with the detail richness quantified by $\sigma$ and $H$ improving by 218.85\% and 56.19\% respectively. In overexposed scenarios, the brightness was adjusted from 245.64 to 150.85, with detail richness up by 187.73\% and 40.65\% respectively. These results confirm the effectiveness of the proposed method in both scenarios, maintaining appropriate brightness and preserving intricate blade details. Suboptimal performance occasionally occurred during rapid sunlight changes. The slight delay in exposure adjustment led to temporary overexposure or underexposure in a few consecutive frames, indicating room to improve response speed for fast-changing lighting.

\renewcommand{\arraystretch}{1.2}
\begin{table}[h]
\centering
\caption{The metrics of the proposed method in two scenarios}
\label{tab:exposure_result}
\begin{tabular}{c c c c} 
\toprule
 & & Original image & Adjusted image  \\
\midrule
\multirow{3}{*}{Underexposure} & Mean $\mu$ & 25.48 & \textbf{133.27} \\
& Std Dev $\sigma$ $\bm{\uparrow}$ & 2.48 & \textbf{7.91} \\
& Entropy $H$ $\bm{\uparrow}$ & 2.10 & \textbf{3.28} \\
\midrule
\multirow{3}{*}{Overexposure} & Mean $\mu$ & 245.64 & \textbf{150.85} \\
& Std Dev $\sigma$ $\bm{\uparrow}$ & 5.46 & \textbf{10.25} \\
& Entropy $H$ $\bm{\uparrow}$ & 1.23 & \textbf{2.96} \\
\bottomrule
\end{tabular}
\end{table}

\section{CONCLUSION AND FUTURE WORK}
\label{sec:conclusion}

In this paper, toward automated UAV-based wind turbine blade inspection, we propose a UAV inspection platform, a Fermat point based blade stop angle estimation approach, and a blade details prioritized exposure adjustment approach. Extensive 120 flights conducted across 10 wind turbine models in 5 operational wind farms have demonstrated their superior performance, with tangible contributions to addressing the core challenges of automated inspection. The platform’s stability ensures consistent data collection in complex wind farm environments, addressing the operational reliability challenges. The blade stop angle estimation approach, with a mean error of 1.15$^\circ$ and a success rate of 98.3\%, enhances blade position accuracy to facilitate more accurate inspection trajectory planning, and reduce flight tracking errors along the blade surface. This is critical for ensuring full coverage of blade inspection. Additionally, the exposure adjustment approach maintains optimal blade brightness and enhances detail clarity, with improvements of 56.19\% and 40.65\% in underexposed and overexposed scenarios. This makes subtle defects distinguishable, directly improving the reliability of downstream damage detection.

In future work, building upon the platform and approaches proposed in this paper, we will aim to conduct more comprehensive wind turbine parameter estimation and inspection flight planning and control, finally achieving a fully automated UAV-based wind turbine blade inspection system.










\bibliographystyle{IEEEtran}
\bibliography{refs}

\end{document}